\documentclass[10pt,twocolumn,letterpaper]{article}

\usepackage{cvpr}
\usepackage{times}
\usepackage{epsfig}
\usepackage{graphicx}
\usepackage{amsmath, bm}
\usepackage{amssymb}
\usepackage{multirow}
\usepackage{microtype}
\usepackage[normalem]{ulem}
\useunder{\uline}{\ul}{}
\usepackage{subfigure}
\usepackage{color}
\usepackage{nicefrac}
\usepackage{comment}
\usepackage{enumitem}
\usepackage{adjustbox}
\usepackage{mathabx}
\usepackage{tabu}
\usepackage{booktabs}
\usepackage{diagbox}
\usepackage{cuted}
\usepackage{capt-of} 
\usepackage[table,x11names]{xcolor}

\definecolor{gray}{RGB}{222,222,222}
\newcommand{\midsepremove}{\aboverulesep = 0mm \belowrulesep = 0mm}
\midsepremove
\newcommand{\midsepdefault}{\aboverulesep = 0.605mm \belowrulesep = 0.984mm}
\midsepdefault

\usepackage{caption}
\usepackage[pagebackref=false,breaklinks=true,letterpaper=true,bookmarks=false]{hyperref}

\cvprfinalcopy 

\def\cvprPaperID{7383} 

\begin{document}

\title{Dreaming to Distill: Data-free Knowledge Transfer via DeepInversion}

\author{
Hongxu Yin$^{1,2\dagger}\thanks{Equal contribution. $^\dagger$ Work done during an internship at NVIDIA. Work supported in part by ONR MURI N00014-16-1-2007.}$ \ , 
Pavlo Molchanov$^{1*}$,  
Zhizhong Li$^{1,3\dagger}$, 
Jose M. Alvarez$^{1}$, \\
Arun Mallya$^{1}$,
Derek Hoiem$^{3}$, 
Niraj K. Jha$^{2}$, 
and Jan Kautz$^{1}$ \\
\vspace{-0.35cm}
\\
$^1$NVIDIA, $^2$Princeton University, $^3$University of Illinois at Urbana-Champaign \\
\tt\small \{hongxuy, jha\}@princeton.edu, 
\tt\small \{zli115, dhoiem\}@illinois.edu, \\
\tt\small \{pmolchanov, josea, amallya, jkautz\}@nvidia.com
}

\maketitle

\begin{strip}\centering

\vspace{-2cm}

\includegraphics[width=1.\textwidth,clip]{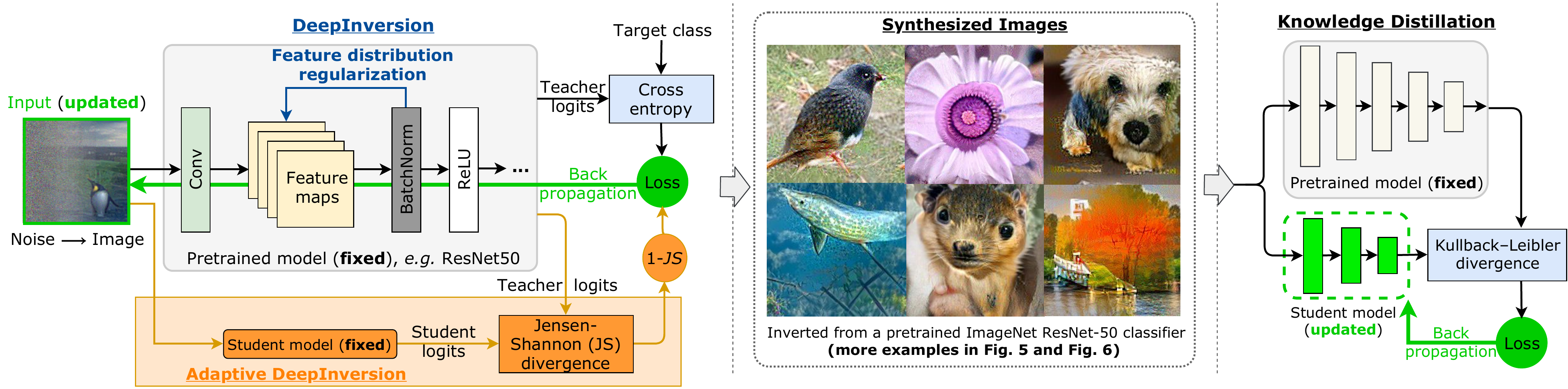}
\captionof{figure}{
We introduce DeepInversion, a method that optimizes random noise into high-fidelity class-conditional images given just a pretrained CNN (teacher), in Sec.~\ref{sec:deepinversion}. 
Further, we introduce Adaptive DeepInversion (Sec.~\ref{sec:adaptivedeepinversion}),
which utilizes both the teacher and application-dependent student network to improve image diversity. Using the synthesized images, we enable data-free pruning (Sec.~\ref{sec:imagenet_pruning}), introduce and address data-free knowledge transfer~(Sec.~\ref{sec:imagenet_teaching}), and improve upon data-free continual learning (Sec.~\ref{sec:imagenet_incremental}).  
\label{fig:teaser}}
\vspace{3mm}
\end{strip}


\begin{abstract}
\vspace{-4mm}
We introduce DeepInversion, a new method for synthesizing images from the image distribution used to train a deep neural network. We ``invert'' a trained network (teacher) to synthesize class-conditional input images starting from random noise, without using any additional information on the training dataset. Keeping the teacher fixed, our method optimizes the input while regularizing the distribution of intermediate feature maps using information stored in the batch normalization layers of the teacher.
Further, we improve the diversity of synthesized images using Adaptive DeepInversion, which maximizes the Jensen-Shannon divergence between the teacher and student network logits. 
The resulting synthesized images from networks trained on the CIFAR-10 and ImageNet datasets demonstrate high fidelity and degree of realism, and help enable a new breed of data-free applications -- ones that do not require any real images or labeled data.
We demonstrate the applicability of our proposed method to three tasks of immense practical importance -- (i) data-free network pruning, (ii) data-free knowledge transfer, and (iii) data-free continual learning. Code is available at \textcolor{magenta}{\url{https://github.com/NVlabs/DeepInversion}}
\end{abstract}

\section{Introduction}

The ability to transfer learned knowledge from a trained neural network to a new one with properties desirable for the task at hand has many appealing applications. For example, one might want to use a more resource-efficient architecture for deployment on edge inference devices~\cite{molchanov2019importance,fbnet,zhu2016trained}, or to adapt the network to the inference hardware~\cite{chamnet, wang2019haq, yin2019hardware}, or for continually learning to classify new image classes~\cite{li2017learning,lopez2015unifying}, \etc. Most current solutions for such knowledge transfer tasks are based on the concept of knowledge distillation~\cite{hinton2015distilling}, wherein the new network (student) is trained to match its outputs to that of a previously trained network (teacher). 
However, all such methods have a significant constraint -- they assume that either the previously used training dataset is available~\cite{chen2015net2net,li2017learning,molchanov2016pruning,romero2014fitnets}, or some real images representative of the prior training dataset distribution are available~\cite{Kimura2018FewshotLO,kimura2018few,lopez2015unifying,rebuffi2017icarl}. Even methods not based on distillation~\cite{Kirkpatrick2016OvercomingCF,nguyen2018variational,Zenke2017ContinualLT} assume that some additional statistics about prior training is made available by the pretrained model provider.

The requirement for prior training information can be very restrictive in practice. For example, suppose a very deep network such as ResNet-152~\cite{he2016deep} was trained on datasets with millions~\cite{deng2009imagenet} or even billions of images~\cite{mahajan2018exploring}, and we wish to distill its knowledge to a lower-latency model such as ResNet-18. In this case, we would need access to these datasets, which are not only large but difficult to store, transfer, and manage. Further, another emerging concern is that of data privacy. While entities might want to share their trained models, sharing the training data might not be desirable due to user privacy, security, proprietary concerns, or competitive disadvantage.

In the absence of prior data or metadata, an interesting question arises -- can we somehow recover training data from the already trained model and use it for knowledge transfer?
A few methods have attempted to visualize what a trained deep network expects to see in an image~\cite{bhardwaj2019dream, mahendran2015understanding, mordvintsev2015deepdream,nguyen2015deep}. The most popular and simple-to-use method is
DeepDream~\cite{mordvintsev2015deepdream}. It synthesizes or transforms an input image to yield high output responses for chosen classes in the output layer of a given classification model. This method optimizes the input (random noise or a natural image), possibly with some regularizers, while keeping the selected output activations fixed, but leaves intermediate representations constraint-free. The resulting ``dreamed" images lack natural image statistics and can be quite easily identified as unnatural. These images are also not very useful for the purposes of transferring knowledge, as our extensive experiments in Section~\ref{sec:all_expts} show.

In this work, we make an important observation about deep networks that are widely used in practice -- they all implicitly encode very rich information about prior training data. 
Almost all high-performing convolutional neural networks (CNNs), such as ResNets~\cite{he2016deep}, DenseNets~\cite{huang2017densely}, or their variants, use the batch normalization layer~\cite{ioffe2015batch}. These layers store running means and variances of the activations at multiple layers. In essence, they store the history of previously seen data, at multiple levels of representation.
By assuming that these intermediate activations follow a Gaussian distribution with mean and variance equal to the running statistics, we show that we can obtain ``dreamed'' images that are realistic and much closer to the distribution of the training dataset as compared to prior work in this area. 

Our approach, visualized in Fig.~\ref{fig:teaser}, called \textit{DeepInversion}, introduces a  regularization term for intermediate layer activations of dreamed images based on just the two layer-wise statistics: mean and variance, which are directly available with trained models. As a result, we do not require any training data or metadata to perform training image synthesis. Our method is able to generate images with high fidelity and realism at a high resolution, as can be seen in the middle section of Fig.~\ref{fig:teaser}, and more samples in Fig.~\ref{fig:imagenet_inversion} and Fig.~\ref{fig:resnetv1.5}.

We also introduce an application-specific extension of \textit{DeepInversion}, called
\textit{Adaptive DeepInversion}, which can enhance the diversity of the generated images. More specifically, it exploits disagreements between the pretrained teacher and the in-training student network to expand the coverage of the training set by the synthesized images. It does so by maximizing the Jensen-Shannon divergence between the responses of the two networks.

In order to show that our dataset synthesis method is useful in practice, 
we demonstrate its effectiveness on three different use cases. First, we show that the generated images support knowledge transfer between two networks using distillation, even with different architectures, with a minimal accuracy loss on the simple CIFAR-10 as well as the large and complex ImageNet dataset. Second, we show that we can prune the teacher network using the synthesized images to obtain a smaller student on the ImageNet dataset. 
Finally, we apply DeepInversion to continual learning that enables the addition of new classes to a pretrained CNN without the need for any original data.
Using our DeepInversion technique, we empower a new class of ``data-free'' applications of immense practical importance, which need neither any natural image nor labeled data.

Our main contributions are as follows:
\begin{itemize}[topsep=1pt,itemsep=1pt,partopsep=0pt, parsep=0pt,leftmargin=\labelwidth]

    \item We introduce DeepInversion, a new method for synthesizing class-conditional images from a CNN trained for image classification (Sec.~\ref{sec:deepinversion}). Further, we improve synthesis diversity by exploiting student-teacher disagreements via Adaptive DeepInversion (Sec.~\ref{sec:adaptivedeepinversion}).

    \item We enable data-free and hardware-aware pruning that achieves performance comparable to the state-of-the-art (SOTA) methods that rely on the training dataset (Sec.~\ref{sec:imagenet_pruning}).  
    
    \item We introduce and address the task of data-free knowledge transfer between a teacher and a randomly initialized student network (Sec.~\ref{sec:imagenet_teaching}). 
    
    \item We improve prior work on data-free continual (a.k.a. incremental) learning, and achieve results comparable to oracle methods given the original data (Sec.~\ref{sec:imagenet_incremental}).
    
\end{itemize}

\section{Related Work}

\noindent\textbf{Knowledge distillation}.
Transfer of knowledge from one model to another was first introduced by Breiman and Shang when they learned a single decision tree to approximate the outputs of multiple decision trees~\cite{breiman1996born}. Similar ideas are explored in neural networks by Bucilua \textit{et al.}~\cite{bucilua2006model}, Ba and Caruana~\cite{ba2014deep}, and Hinton \textit{et al.}~\cite{hinton2015distilling}. Hinton \etal formulate the problem as ``knowledge distillation,'' where a compact student mimics the output distributions of expert teacher models~\cite{hinton2015distilling}. These methods and improved variants~\cite{ahn2019variational,park2019relational,romero2014fitnets,xu2017training,zagoruyko2016paying} enable teaching students with goals such as quantization~\cite{ mishra2017apprentice,polino2018model}, compact neural network architecture design~\cite{romero2014fitnets}, semantic segmentation~\cite{liu2019structured}, self-distillation~\cite{furlanello2018born}, and un-/semi-supervised learning~\cite{lopez2015unifying, pilzer2019refine, yim2017gift}. All these methods still rely on images from the original or proxy datasets. 
More recent research has explored data-free knowledge distillation. Lopes \etal~\cite{lopes2017data} synthesize inputs based on pre-stored auxiliary layer-wise statistics of the teacher network. Chen \textit{et al.}~\cite{chen2019data} train a new generator network for image generation while treating the teacher network as a fixed discriminator. Despite remarkable insights, scaling to tasks such as ImageNet classification, remains difficult for these methods.

\noindent\textbf{Image synthesis.} 
Generative adversarial networks (GANs)~\cite{gulrajani2017improved,miyato2018spectral,nguyen2017plug,zhang2018self} have been at the forefront of generative image modeling, yielding high-fidelity images, \eg, using BigGAN~\cite{brock2018biggan}. Though adept at capturing the image distribution, training a GAN's generator requires access to the original data. 

An alternative line of work in security focuses on image synthesis from a single CNN. Fredrikson \etal~\cite{fredrikson2015modelinversionattack} propose the \textit{model inversion} attack to obtain class images from a network through a gradient descent on the input. Follow-up works have improved or expanded the approach to new threat scenarios~\cite{he2019model, wang2015regression,yang2019adversarial}. These methods have only been demonstrated on shallow networks, or require extra information (\eg, intermediate features). 

In vision, researchers visualize neural networks to understand their properties. Mahendran \etal explore inversion, activation maximization, and caricaturization to synthesize ``natural pre-images'' from a trained network
~\cite{mahendran2015understanding,mahendran2016visualizing}. Nguyen \etal use a trained GAN's generator as a prior to invert trained CNNs~\cite{nguyen2016synthesizing} to images, and its followup Plug \& Play~\cite{nguyen2017plug} further improves image diversity and quality via latent code prior. Bhardwaj \etal use the training data cluster centroids to improve inversion~\cite{bhardwaj2019dream}. These methods still rely on auxiliary dataset information or additional pretrained networks. Of particular relevance to this work is DeepDream~\cite{mordvintsev2015deepdream} by Mordvintsev \etal, which has enabled the ``dreaming'' of new object features onto natural images given a single pretrained CNN. Despite notable progress, synthesizing high-fidelity and high-resolution natural images from a deep network remains challenging.

\section{Method}
Our new data-free knowledge distillation framework consists of two steps: (i) model inversion, and (ii) application-specific knowledge distillation.
In this section, we briefly discuss the background and notation, and then introduce our \textit{DeepInversion} and \textit{Adaptive DeepInversion} methods.

\subsection{Background}
\noindent{\bf Knowledge distillation.}
Distillation~\cite{hinton2015distilling} is a popular technique for knowledge transfer between two models. In its simplest form, first, the teacher, a large model or ensemble of models, is trained. Second, a smaller model, the student, is trained to mimic the behavior of the teacher by matching the temperature-scaled soft target distribution produced by the teacher on training images (or on other images from the same domain). 
Given a trained model $p_{T}$ and a dataset $\mathcal{X}$, the parameters of the student model, $\textbf{W}_S$, can be learned by 
\begin{equation}
\label{eqn:kd}
   \underset{\textbf{W}_S}{\text{min }} \sum_{x \in \mathcal{X}}\text{KL} (p_{T}(x), p_{S}(x)),
\end{equation}
where KL$(\cdot)$ refers to the Kullback-Leibler divergence and $p_T(x) = p(x, \textbf{W}_T)$ and $p_S(x) = p(x, \textbf{W}_S)$ are the output distributions produced by the teacher and student model, respectively, typically obtained using a high temperature on the softmax inputs~\cite{hinton2015distilling}.

Note that ground truths are not required. 
Despite its efficacy, the process still relies on real images from the same domain.
Below, we focus on methods to synthesize a large set of images $\hat{x} \in \hat{\mathcal{X}}$ from noise that could replace $x \in \mathcal{X}$. 
\medskip

\noindent{\bf DeepDream~\cite{mordvintsev2015deepdream}.} Originally formulated by Mordvintsev \etal to derive artistic effects on natural images, DeepDream is also suitable for optimizing noise into images. 
Given a randomly initialized input ($\hat{x}\in \mathcal{R}^{H\times W\times C}$, $H, W, C$ being the height, width, and number of color channels) and an arbitrary target label $y$, the image is synthesized by optimizing
\begin{equation}
\underset{\hat{x}}{\text{min }}{ {\mathcal{L}(\hat{x},y) +  \mathcal{R}(\hat{x})}}, 
\label{eqn:main_error}
\end{equation}
where $\mathcal{L(\cdot)}$ is a classification loss (e.g., cross-entropy), and $\mathcal{R(\cdot)}$ is an image regularization term. 
DeepDream uses an image prior~\cite{dosovitskiy2016inverting,mahendran2015understanding,nguyen2015deep,simonyan2013deep} to steer $\hat x$ away from unrealistic images with no discernible visual information:
\begin{equation}
    \mathcal{R}_{\text{prior}}(\hat{x}) = \alpha_{\text{tv}} \mathcal{R}_{\text{TV}}(\hat{x}) + \alpha_{\ell_{2}} \mathcal{R}_{\ell_2}(\hat{x}),
\label{eqn:r_prior}
\end{equation}
where $R_{\text{TV}}$ and $R_{\ell_2}$ penalize the total variance and $\ell_2$ norm of $\hat{x}$, respectively, with scaling factors $\alpha_{\text{tv}}$, $\alpha_{\ell_{2}}$. As both prior work~\cite{mahendran2015understanding, mordvintsev2015deepdream, nguyen2015deep} and we empirically observe, 
image prior regularization provides more stable convergence to valid images. However, these images still have a distribution far different from natural (or original training) images and thus lead to unsatisfactory knowledge distillation results. 

\subsection{DeepInversion (DI)} 
\label{sec:deepinversion}
We improve DeepDream's image quality by extending image regularization $\mathcal{R}(\hat{x})$ with a new feature distribution regularization term. The image prior term defined previously provides little guidance for obtaining a synthetic $\hat{x}\in \hat{\mathcal{X}}$ that contains similar low- and high-level features as $x\in\mathcal{X}$. To effectively enforce feature similarities at all levels, we propose to minimize the distance between feature map statistics for $\hat{x}$ and $x$. 
We assume that feature statistics follow the Gaussian distribution across batches and, therefore, can be defined by mean $\mu$ and variance $\sigma^2$. Then, the \textit{feature distribution regularization} term can be formulated as:
\begin{eqnarray}
  \begin{aligned}
    \mathcal{R}_{\text{feature}}(\hat{x}) 
    = & \sum_{l}|| \ \mu_{l}(\hat{x}) - \mathbb{E}(\mu_{l}({x}) | \mathcal{X}) \ ||_2 + \\ &\sum_{l}|| \ {\sigma^2_l}(\hat{x}) - \mathbb{E}({\sigma^2_{l}(x) | \mathcal{X}) \ ||_2},
\end{aligned}
\label{eqn:fmap}
\end{eqnarray}
where $\mu_{l}(\hat{x})$ and $\sigma_l^2(\hat{x})$ are the batch-wise mean and variance estimates of feature maps corresponding to the $l^{\text{th}}$ convolutional layer. The $\mathbb{E}(\cdot)$ and $||\cdot||_2$ operators denote the expected value and $\ell_2$ norm calculations, respectively. 

It might seem as though a set of training images would be required to obtain $\mathbb{E}(\mu_{l}(x) | \mathcal{X})$ and $\mathbb{E}(\sigma^2_{l}(x) | \mathcal{X})$, but the running average statistics stored in the widely-used BatchNorm (BN) layers are more than sufficient. A BN layer normalizes the feature maps during training to alleviate covariate shifts~\cite{ioffe2015batch}. It implicitly captures the channel-wise means and variances during training, hence allows for estimation of the expectations in Eq.~\ref{eqn:fmap} by:
\begin{equation}
    \mathbb{E}\big(\mu_{l} (x)| \mathcal{X}\big) \simeq \text{BN}_{l}( \text{running\_mean} ),
\end{equation}
\begin{equation}
    \mathbb{E}\big(\sigma^2_{l} (x)| \mathcal{X}\big) \simeq \text{BN}_{l}(\text{running\_variance}).
\end{equation}

As we will show, this feature distribution regularization substantially improves the quality of the generated images. We refer to this model inversion method as \textit{DeepInversion} $-$ a generic approach that can be applied to any trained \textit{deep} CNN classifier for the \textit{inversion} of high-fidelity images. $R(\cdot)$ (corr. to Eq.~\ref{eqn:main_error}) can thus be expressed as
\begin{equation}
\mathcal{R}_{\text{DI}}(\hat{x}) = \mathcal{R}_{\text{prior}}(\hat{x}) + \alpha_{\text{f}} \mathcal{R}_{\text{feature}}(\hat{x}).
\label{eqn:di}
\end{equation}

\subsection{Adaptive DeepInversion (ADI)}
\label{sec:adaptivedeepinversion}
In addition to quality, diversity also plays a crucial role in avoiding repeated and redundant synthetic images. Prior work on GANs has proposed various techniques, such as min-max training competition~\cite{goodfellow2014generative} and the truncation trick~\cite{brock2018biggan}. These methods rely on the joint training of two networks over original data and therefore are not applicable to our problem. We propose \textit{Adaptive DeepInversion}, an enhanced image generation scheme based on a novel iterative competition scheme between the image generation process and the student network. The main idea is to encourage the synthesized images to cause student-teacher disagreement. For this purpose, we introduce an additional loss $\mathcal{R}_{\text{compete}}$ for image generation based on the Jensen-Shannon divergence that penalizes output distribution similarities,
\begin{eqnarray}
\mathcal{R}_\text{compete}(\hat{x})&=& 1- \text{JS}(p_{T}(\hat{x}),p_{S}(\hat{x})), 
\label{eq:competition}
\end{eqnarray}\vspace{-9mm}
\begin{eqnarray}
\text{JS}(p_{T}(\hat{x}),p_{S}(\hat{x}))={\displaystyle  \frac{1}{2}\bigg(\text{KL}(p_{T}(\hat{x}),M)+\text{KL}(p_{S}(\hat{x}),M)\bigg)}, \nonumber
\end{eqnarray}
where $M = \frac{1}{2}\cdot\big(p_{T}(\hat{x})+p_{S}(\hat{x})\big)$ is the average of the teacher and student distributions. 

\begin{figure}[t]
\begin{center}
\includegraphics[width=\columnwidth]{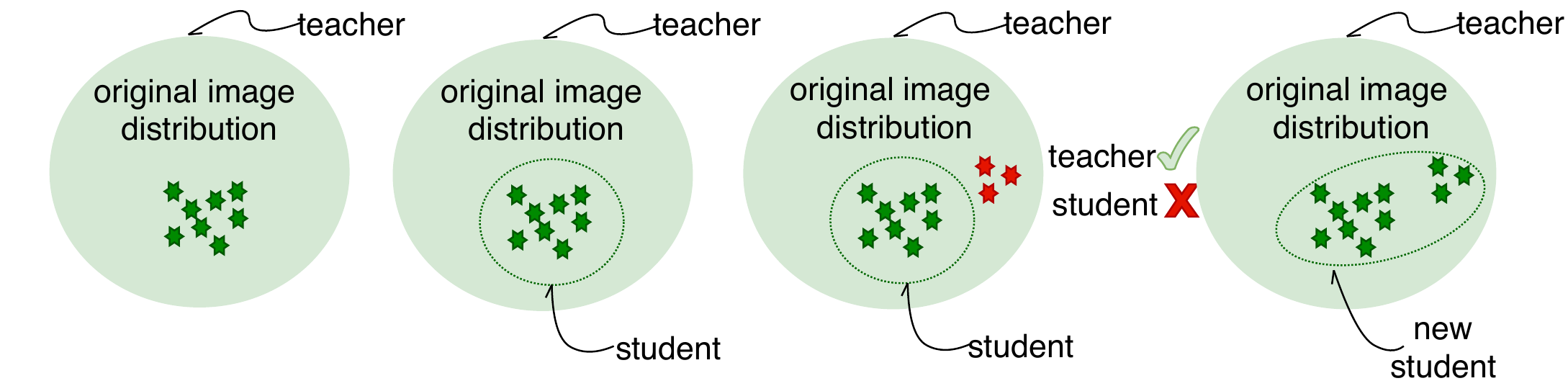}
\end{center}
\vskip-2mm
\caption{Illustration of the Adaptive DeepInversion competition scheme to improve image diversity. Given a set of generated images (shown as green stars), an intermediate student can learn to capture part of the original image distribution. Upon generating new images (shown as red stars), competition encourages new samples out of student's learned knowledge, improving distributional coverage and facilitating additional knowledge transfer. Best viewed in color.
}
\label{fig:competition}
\end{figure}

During optimization, this new term leads to new images the student cannot easily classify whereas the teacher can. 
As illustrated in Fig.~\ref{fig:competition}, our proposal iteratively expands the distributional coverage of the image distribution during the learning process. 
With competition, regularization $R(\cdot)$ from Eq.~\ref{eqn:di} is updated with an additional loss scaled by $\alpha_{c}$ as
\begin{equation}
\mathcal{R}_{\text{ADI}}(\hat{x}) =
\mathcal{R}_{\text{DI}}(\hat{x}) + \alpha_{c} \mathcal{R}_{\text{compete}}(\hat{x}).
\end{equation}
\vspace{-5mm}

\subsection{DeepInversion vs.\ Adaptive DeepInversion}

DeepInversion is a generic method that can be applied to any trained CNN classifier. For knowledge distillation, it enables a one-time synthesis of a large number of images given the teacher, to initiate knowledge transfer. Adaptive DeepInversion, on the other hand, needs a student in the loop to enhance image diversity. Its competitive and interactive nature favors constantly-evolving students, which gradually force new image features to emerge, and enables the augmentation of DeepInversion, as shown in our experiments.

\section{Experiments}
\label{sec:all_expts}
We demonstrate our inversion methods on datasets of increasing size and complexity. 
We perform a number of ablations to evaluate each component in our method on the simple CIFAR-10 dataset ($32\times 32$ pixels, 10 classes). Then, on the complex ImageNet dataset ($224\times 224$ pixels, 1000 classes), we show the success of our inversion methods on three different applications under the data-free setting: (a) pruning, (b) knowledge transfer, and (c) continual (class-incremental) learning. In all experiments, image pixels are initialized $i.i.d.$ from Gaussian noise of $\mu=0$ and $\sigma=1$. 

\subsection{Results on CIFAR-10}
\label{sec:cifar_results}

For validating our design choices, we consider the task of data-free knowledge distillation, where we teach a student network randomly initialized from scratch.

\noindent\textbf{Implementation details.}
We use VGG-11-BN and ResNet-34 networks pretrained on CIFAR-10 as the teachers. For all image synthesis in this section, we use Adam for optimization (learning rate 0.05). We generate $32\times32$ images in batches of 256. Each image batch requires 2k gradient updates. After a simple grid search optimizing for student accuracy, we found $\alpha_{\text{tv}}=2.5\cdot 10^{-5}, \alpha_{\ell_2}=3\cdot10^{-8}$, and $\alpha_{f}=\{1.0, 5.0, 10.0, 100.0\}$ work best for DeepInversion, and $\alpha_{c}=10.0$ for Adaptive DeepInversion. See supplementary materials for more details.  

\noindent\textbf{Baselines -- Noise \& DeepDream~\cite{mordvintsev2015deepdream}.}
From Table~\ref{tab:cifar_prune}, we observe that optimized noise, Noise ($\mathcal{L}$), does not provide any support for knowledge distillation $-$ a drastic change in input distribution disrupts the teacher and impacts the validity of the transferred knowledge. Adding $R_{\text{prior}}$, like in DeepDream, slightly improves the student's accuracy.

\begin{figure}[t]
\centering
\resizebox{0.99\linewidth}{!}{
\begingroup
\renewcommand*{\arraystretch}{0.3}
\begin{tabular}{ccc}
\includegraphics[width=0.16\linewidth,clip,trim=5px 0 0 4px]{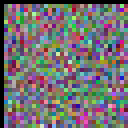} 
\includegraphics[width=0.16\linewidth,clip,trim=5px 0 0 4px]{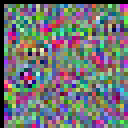} &
\includegraphics[width=0.16\linewidth,clip,trim=5px 0 0 4px]{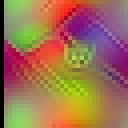} 
\includegraphics[width=0.16\linewidth,clip,trim=5px 0 0 4px]{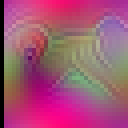} &
\includegraphics[width=0.16\linewidth,clip,trim=5px 0 0 4px]{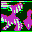} 
\includegraphics[width=0.16\linewidth,clip,trim=5px 0 0 4px]{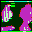} 
\\
\includegraphics[width=0.16\linewidth,clip,trim=5px 0 0 4px]{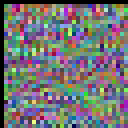} 
\includegraphics[width=0.16\linewidth,clip,trim=5px 0 0 4px]{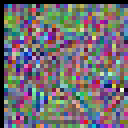} &
\includegraphics[width=0.16\linewidth,clip,trim=5px 0 0 4px]{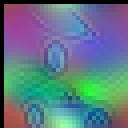} 
\includegraphics[width=0.16\linewidth,clip,trim=5px 0 0 4px]{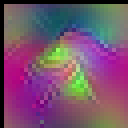} &
\includegraphics[width=0.16\linewidth,clip,trim=5px 0 0 4px]{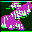} 
\includegraphics[width=0.16\linewidth,clip,trim=5px 0 0 4px]{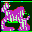} 
\\
(a) Noise (opt) & (b) DeepDream~\cite{mordvintsev2015deepdream} & (c) DAFL~\cite{chen2019data} \\
\\
\end{tabular}
\endgroup
}

\resizebox{0.66\linewidth}{!}{
\begingroup
\renewcommand*{\arraystretch}{0.3}
\begin{tabular}{cc}
\includegraphics[width=0.16\linewidth,clip,trim=5px 0 0 4px]{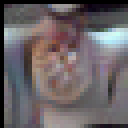}
\includegraphics[width=0.16\linewidth,clip,trim=5px 0 0 4px]{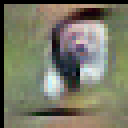} &
\includegraphics[width=0.16\linewidth,clip,trim=4px 0 0 4px]{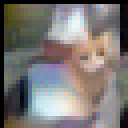}
\includegraphics[width=0.16\linewidth,clip,trim=4px 0 0 4px]{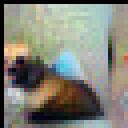} 
\\
\includegraphics[width=0.16\linewidth,clip,trim=5px 0 0 4px]{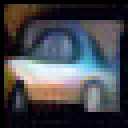} 
\includegraphics[width=0.16\linewidth,clip,trim=5px 0 0 4px]{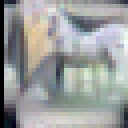} &
\includegraphics[width=0.16\linewidth,clip,trim=5px 0 0 4px]{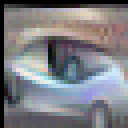} 
\includegraphics[width=0.16\linewidth,clip,trim=5px 0 0 4px]{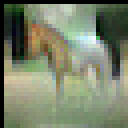} 
\\
(d) DeepInversion (DI) & (e) Adaptive DI (ADI) \\
\end{tabular}
\endgroup
}\vskip1mm
\caption{$32\times32$ images generated by inverting a ResNet-34 trained on CIFAR-10 with different methods. All images are correctly classified by the network, clockwise: cat, dog, horse, car.
}
\vspace{-3mm}
\label{fig:cifar_inversion_ablation}
\end{figure}

\noindent\textbf{Effectiveness of  DeepInversion ($R_{\text{feature}}$).} Upon adding $R_{\text{feature}}$, we immediately find large improvements in accuracy of $40\%$-$69\%$ across all the teaching scenarios. DeepInversion images (Fig.~\ref{fig:cifar_inversion_ablation}(d)) are vastly superior in realism, as compared to the baselines (Fig.~\ref{fig:cifar_inversion_ablation}(a,b)).

\begin{figure}[t!]
\centering
\includegraphics[width=0.98\linewidth,trim={0 0.3cm 0 0}]{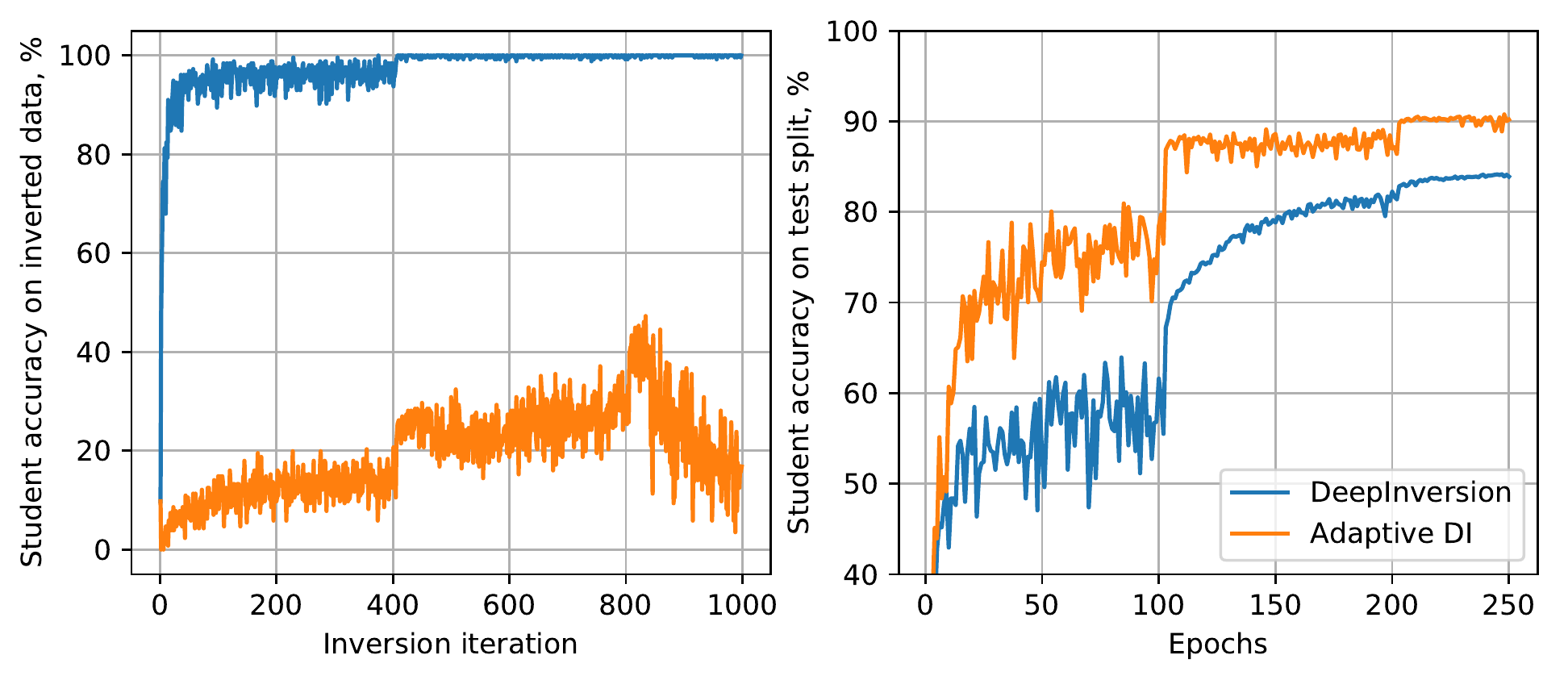}
\caption{Progress of knowledge transfer from trained VGG-11-BN (92.34\% acc.) to freshly initialized VGG-11-BN network (student) using inverted images. Plotted are accuracies on generated (left) and real (right) images. Final student accuracies shown in Table~\ref{tab:cifar_prune}. }
\label{fig:cifar10_training}
\vspace{-2mm}
\end{figure}

\noindent\textbf{Effectiveness of Adaptive DeepInversion ($R_{\text{compete}}$).} Using competition-based inversion further improves accuracy by $1\%$-$10\%$, bringing the student accuracy very close to that of the teacher trained on real images from the CIFAR-10 dataset (within $2\%$). 
The training curves from one of the runs are shown in Fig.~\ref{fig:cifar10_training}. 
Encouraging teacher-student disagreements brings in additional ``harder'' images during training (shown in Fig.~\ref{fig:cifar_inversion_ablation}(e)) that remain correct for the teacher, but have a low student accuracy, as can be seen from Fig.~\ref{fig:cifar10_training}~(left). 

\noindent\textbf{Comparison with DAFL~\cite{chen2019data}.} We further compare our method with DAFL~\cite{chen2019data}, which trains a new generator network to convert noise into images using a fixed teacher.
As can be seen from Fig.~\ref{fig:cifar_inversion_ablation}(c), we notice that these images are ``unrecognizable,'' reminiscent of ``fooling images''~\cite{nguyen2015deep}.
Our method enables higher visual fidelity of images and eliminates the need for an additional generator network, while gaining higher student accuracy under the same setup.

\begin{table}[!t]
\centering
\resizebox{.99\linewidth}{!}{
\begin{tabular}{lccc}
\toprule

Teacher Network &    VGG-11 & VGG-11 & ResNet-34 \\
Student Network &  VGG-11 & ResNet-18 & ResNet-18 \\
Teacher accuracy                 & 92.34\%    & 92.34\%  & 95.42\%    \\
\midrule 
Noise ($\mathcal{L}$)                & 13.55\%  &  13.45\% & 13.61\%  \\
$+\mathcal{R}_\text{prior}$ (DeepDream~\cite{mordvintsev2015deepdream})                  & 36.59\% & 39.67\%    & 29.98\%              \\
$+\mathcal{R}_\text{feature}$ (DeepInversion)                   & 84.16\%      & 83.82\%   &  91.43\%     \\
$+\mathcal{R}_\text{compete}$ (ADI)         & \textbf{90.78\%}  & \textbf{90.36\%} &  \textbf{93.26\%}\\
\midrule 
DAFL~\cite{chen2019data} &  -- & -- &92.22\% \\
\bottomrule

\end{tabular}
}
\caption{Data-free knowledge transfer to various students on CIFAR-10. For ADI, we generate one new batch of images every 50 knowledge distillation iterations and merge the newly generated images into the existing set of generated images. }
\label{tab:cifar_prune}

\end{table}

\subsection{Results on ImageNet}
\label{sec:imagenet_deepinversion}
After successfully demonstrating our method's abilities on the small CIFAR-10 dataset, we move on to examine its effectiveness on the large-scale ImageNet dataset~\cite{deng2009imagenet}.
We first run DeepInversion on networks trained on ImageNet, and perform quantitative and qualitative analyses.
Then, we show the effectiveness of synthesized images on three different tasks of immense importance: data-free pruning, data-free knowledge transfer, and data-free continual learning.

\begin{figure*}[!t]
\centering
\includegraphics[width=1.\textwidth]{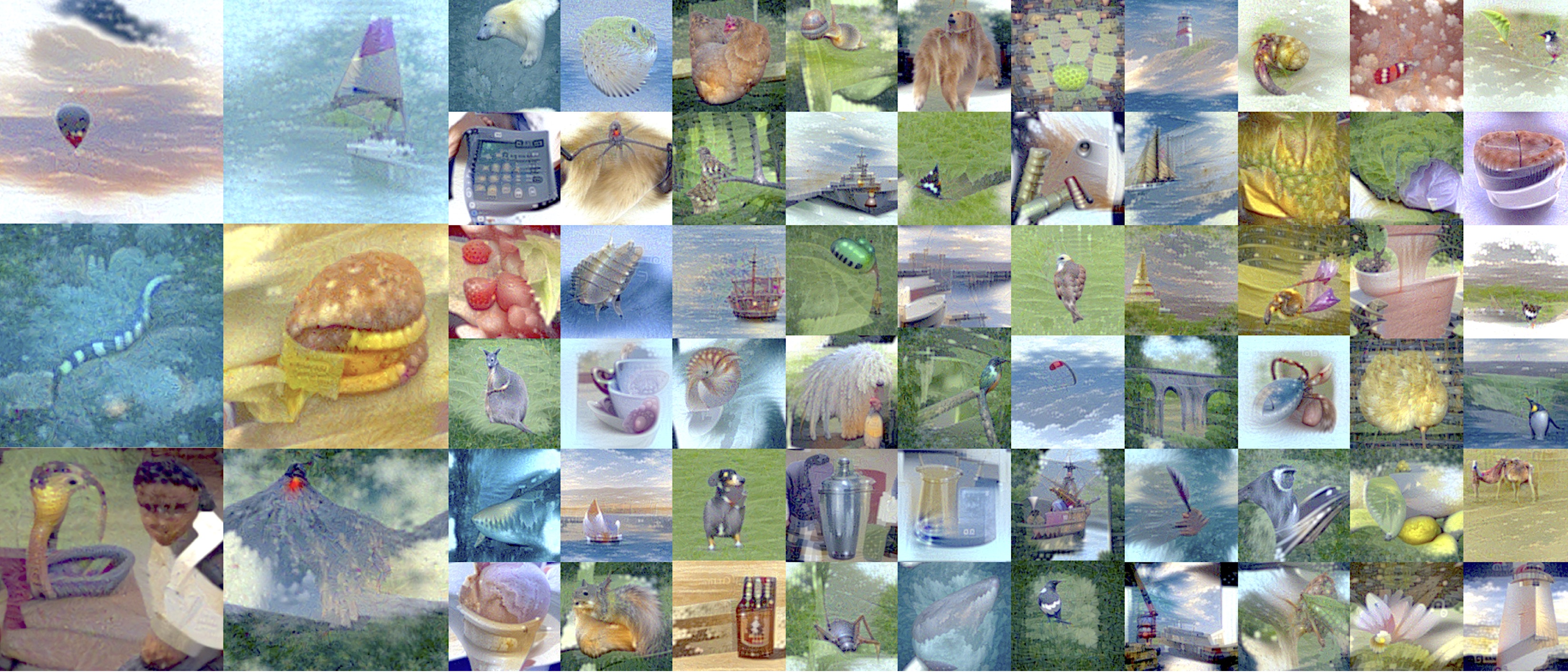}
\caption{Class-conditional $224\times224$ samples obtained by DeepInversion, given only a ResNet-50 classifier trained on ImageNet and no additional information. Note that the images depict classes in contextually correct backgrounds, in realistic scenarios.
Best viewed in color.}
\label{fig:imagenet_inversion}
\end{figure*}

\noindent\textbf{Implementation details.}
For all experiments in this section, we use the publicly available pretrained ResNet-\{18,~50\} from PyTorch as the fixed teacher network, with top-1 accuracy of \{$69.8\%,76.1\%$\}. For image synthesis, we use Adam for optimization (learning rate $0.05$). We set $\alpha_{\text{tv}}=1\cdot10^{-4}, \alpha_{\ell_2}=\{0, 1\cdot10^{-2}\}, \alpha_{f}=1\cdot10^{-2}$ for DeepInversion, and $\alpha_{c}=0.2$ for Adaptive DeepInversion. We synthesize $224\times224$ images in batches of $1,216$ using 8 NVIDIA V100 GPUs and automatic-mixed precision (AMP)~\cite{micikevicius2017mixed} acceleration. Each image batch receives 20k updates over 2h. 
\vspace{-3mm}

\subsubsection{Analysis of DeepInversion Images}
Fig.~\ref{fig:imagenet_inversion} shows images generated by DeepInversion from an ImageNet-pretrained ResNet-50. 
Remarkably, given just the model, we observe that DeepInversion is able to generate images with high fidelity and resolution.
It also produces detailed image features and textures around the target object, \eg,  clouds surrounding the target balloon, water around a catamaran, forest below the volcano, \etc. 

\noindent\textbf{Generalizability.} 
In order to verify that the generated images do not overfit to just the inverted model, we obtain predictions using four other ImageNet networks.
As can be seen from Table~\ref{tab:verifier_acc}, images generated using a ResNet-50 generalize to a range of models and are correctly classified. Further, DeepInversion
outperforms DeepDream by a large margin. This indicates robustness of our generated images while being transferred across networks.

\noindent\textbf{Inception score (IS).} 
We also compare the IS~\cite{salimans2016improved} of our generated images with other methods in Table~\ref{tab:imagenet_is_score}. Again, DeepInversion substantially outperforms DeepDream with an improvement of $54.2$. Without sophisticated training, DeepInversion even beats multiple GAN baselines that have limited scalability to high image resolutions.

\subsection{Application I: Data-free Pruning}
\label{sec:imagenet_pruning}
Pruning removes individual weights or entire filters (neurons) from a network such that the metric of interest (\eg, accuracy, precision) does not drop significantly.
Many techniques have been proposed to successfully compress neural networks~\cite{han2015deep, li2016pruning, liu2017learning, thinet, molchanov2019importance, molchanov2016pruning, ye2018rethinking, nisp}. All these methods  require images from the original dataset to perform pruning. We build upon the pruning method of Molchanov \etal~\cite{molchanov2019importance}, which uses the Taylor approximation of the pruning loss for a global ranking of filter importance over all the layers. 
We extend this method by applying the KL divergence loss, computed between the teacher and student output distributions. Filter importance is estimated from images inverted with DeepInversion and/or Adaptive DeepInversion, making pruning data-free. We follow the pruning and finetuning (30 epochs) setup from~\cite{molchanov2019importance}. All experiments on pruning are performed with ResNet-50.

\begin{table}[!t]
\centering
\resizebox{.75\linewidth}{!}{
\begin{tabular}{lcc}
\toprule
        \multirow{2}{*}{Model}               & DeepDream       & DeepInversion \\  
                            & top-1 acc. (\%)  & top-1 acc. (\%)    \\
                            \midrule
    ResNet-50               & $100$           &       100   \\ \midrule
    ResNet-18               & $28.0$          &       $\bf{94.4}$   \\
    Inception-V3            & $27.6$          &       $\bf{92.7}$   \\
    MobileNet-V2            & $13.9$          &       $\bf{90.9}$   \\
    VGG-11                  & $6.7$           &       $\bf{80.1}$   \\
    \bottomrule
\end{tabular}
}
\caption{Classification accuracy of ResNet-50 synthesized images by other ImageNet-trained CNNs.}
\label{tab:verifier_acc}
\end{table}
\begin{table}[!t]
\centering
\resizebox{.93\linewidth}{!}{
\begin{tabular}{lc c c}
\toprule
    Method &  Resolution               & GAN         & Inception Score    \\\midrule
    BigGAN~\cite{brock2018biggan} & $256$   & \checkmark  & $178.0$ / $202.6^+$ \\
    DeepInversion (\textbf{Ours}) & $224$ &              & $60.6$  \\
    SAGAN~\cite{zhang2018self}  & $128$                 & \checkmark  & $52.5$  \\
    SNGAN~\cite{miyato2018spectral}  & $128$                 & \checkmark   & $35.3$  \\
    WGAN-GP~\cite{gulrajani2017improved} & $128$                & \checkmark   & $11.6$  \\
    DeepDream~\cite{mordvintsev2015deepdream}* & $224$              &    & $6.2$  \\\bottomrule
\end{tabular}
}
\caption{Inception Score (IS) obtained by images synthesized by various methods on ImageNet. SNGAN ImageNet score from~\cite{shmelkov2018good}. *: our implementation.  $^+$: BigGAN-deep.}
\label{tab:imagenet_is_score}
\end{table}

\noindent\textbf{Hardware-aware loss.} We further  consider actual latency on the target hardware for a more efficient pruning. To achieve this goal, the importance ranking of filters needs to reflect not only accuracy but also latency, quantified by: %
\begin{equation}
    \mathcal{I}_\mathcal{S}  (\textbf{W}) = \mathcal{I}_{\mathcal{S}, err}  (\textbf{W}) + \eta \ \mathcal{I}_{\mathcal{S}, lat}  (\textbf{W}),
\label{eq:latency_aware_loss}
\end{equation}
where $\mathcal{I}_{\mathcal{S}, err}$ and $ \mathcal{I}_{\mathcal{S}, lat}$, respectively, focus on absolute changes in network error and inference latency, specifically, when the filter group $s\in\mathcal{S}$ is zeroed out from the set of neural network parameters $\textbf{W}$. $\eta$ balances their importance. We approximate the latency reduction term, $\mathcal{I}_{\mathcal{S}, lat}$, via precomputed hardware-aware look-up tables of operation costs (details in the Appendix). 

\noindent\textbf{Data-free pruning evaluation.}
For better insights, we study four image sources: \textbf{(i)} partial ImageNet with 0.1M original images; \textbf{(ii)} unlabeled images from the proxy dataset, MS COCO~\cite{lin2014microsoft} (127k images), and PASCAL VOC~\cite{everingham2010pascal} (9.9k images) datasets; \textbf{(iii)} 100k generated images from the BigGAN-deep~\cite{brock2018biggan} model, and \textbf{(iv)} a data-free setup with the proposed methods: we first generate 165k images via DeepInversion, and then add to the set an additional 24k/26k images through two competition rounds of Adaptive DeepInversion, given pruned students at $61.9\%$/$73.0\%$ top-1 acc. The visualization of the diversity increase due to competition loss (Eq.~\ref{eq:competition}) in Adaptive DeepInversion is shown in Section~\ref{sup:distribution} of the Appendix. 

Results of pruning and fine-tuning are summarized in Table~\ref{tab:imagenet_dataset_compare}. 
Our approach boosts the top-1 accuracy by more than $54\%$ given inverted images. Adaptive DeepInversion performs relatively on par with BigGAN. Despite beating VOC, we still observe a gap between synthesized images (Adaptive DeepInversion and BigGAN) and natural images (MS COCO and ImageNet), which narrows as fewer filters are pruned.

\begin{table}[!t]
\centering
\resizebox{0.88\linewidth}{!}{
\midsepremove
\begin{tabular}{lcc}
\toprule
 \multirow{3}{*}{Image Source}                &\multicolumn{2}{c}{Top-1 acc. (\%) } \\ 
                          & $-50\%$ filters      & $-20\%$ filters \\
                          & $-71\%$ FLOPs        & $-37\%$ FLOPs   \\ \midrule
No finetune             & \textit{$1.9$}                              & \textit{$16.6$}       \\     

\midrule
\rowcolor{gray}\multicolumn{3}{c}{Partial ImageNet} \\
\rowcolor{white} 
0.1M images / 0 label                  & $69.8$                              & $74.9$       \\ 
\midrule
\rowcolor{gray}\multicolumn{3}{c}{Proxy datasets} \\
\rowcolor{white} 
MS COCO                   & $66.0$                              & $73.8$       \\
PASCAL VOC                & $54.4$                              & $70.8$       \\ 
\midrule
\rowcolor{gray}\multicolumn{3}{c}{GAN} \\
\rowcolor{white} Generator, BigGAN        & $63.0$                              & $73.7$       \\ 
\midrule 
\rowcolor{gray}\multicolumn{3}{c}{Noise ({\bf Ours})} \\
\rowcolor{white} 
DeepInversion (DI)                & $55.9$                              & $72.0$       \\              
Adaptive DeepInversion (ADI)       & $60.7$                              & $73.3$       \\ 
\bottomrule                                                   
\end{tabular}
}
\midsepdefault
\caption{ImageNet ResNet-50 pruning results for the knowledge distillation setup, given different types of input images. }
\vspace{4mm}
\label{tab:imagenet_dataset_compare}
\end{table}
\begin{table}[]
\centering
\setlength{\tabcolsep}{4pt}
\midsepremove
\resizebox{1.\linewidth}{!}{
\begin{tabular}{lcccc}
\toprule
\multirow{1}{*}{Method}                     &   ImageNet data  & GFLOPs   &  Lat. (ms)  & Top-1 acc. (\%)  
\\
\midrule
Base model                                      &  \checkmark   & $4.1$       & $4.90$      & $76.1$ \\
\midrule
Taylor~\cite{molchanov2019importance}           & \checkmark    & $2.7$ ($1.5\times$)       & $4.38$ ($1.1\times$)    & $75.5$ \\
SSS~\cite{huang2017data}                        & \checkmark    & $2.8$ ($1.5\times$)      & -         & $74.2$ \\
ThiNet-70~\cite{thinet}                         & \checkmark    & $2.6$ ($1.6\times$)      & -         & $72.0$ \\
NISP-50-A~\cite{nisp}                           & \checkmark    & $3.0$ ($1.4\times$)      & -         & $72.8$ \\
\midrule
\rowcolor{gray}\multicolumn{5}{c}{\textbf{Ours}} \\
Hardware-Aware (HA)                             & \checkmark    & $3.1$ ($1.3\times$)      & $4.24$ (1.2$\times$)     & $76.1$ \\ 
ADI (Data-free)                                 &               & $2.7$ ($1.5\times$)      & $4.36$ (1.1$\times$)     & $73.3$ \\ 
ADI + HA                                        &               & $2.9$ ($1.4\times$)      & $4.22$ (1.2$\times$)     & $74.0$ \\ 
\bottomrule
\end{tabular}
}
\midsepdefault
\caption{ImageNet ResNet-50 pruning comparison with prior work.}
\label{tab:against_sota}
\end{table}

\noindent\textbf{Comparisons with SOTA.}  We compare data-free pruning against SOTA methods in Table~\ref{tab:against_sota} for the setting in which $20\%$ of filters are pruned away globally. 
We evaluate three setups for our approach: (i) individually applying the hardware-aware technique (HA), (ii) data-free pruning with DeepInversion and Adaptive DeepInversion (ADI), and (iii) jointly applying both (ADI+HA). 
First, we evaluate the hardware-aware loss on the original dataset, and achieve a $16\%$ faster inference with zero accuracy loss compared to the base model, we also observe  improvements in inference speed and accuracy over the pruning baseline~\cite{molchanov2019importance}.
In a data-free setup, we achieve a similar post-pruned model performance compared to prior works (which use the original dataset), while completely removing the need for any images/labels. The data-free setup demonstrates a $2.8\%$ loss in accuracy with respect to the base model. A final combination of both data-free and hardware-aware techniques (ADI+HA) closes this gap to only $2.1\%$, with faster inference.

\subsection{Application II: Data-free Knowledge Transfer}
\label{sec:imagenet_teaching}
\begin{figure*}
\centering
\includegraphics[width=0.88\textwidth]{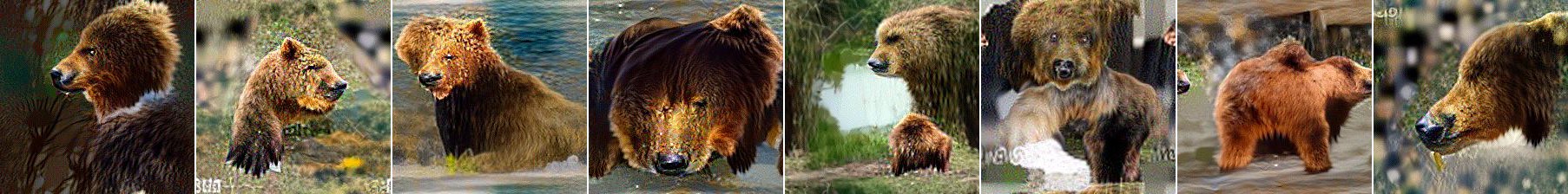}
\includegraphics[width=0.88\textwidth]{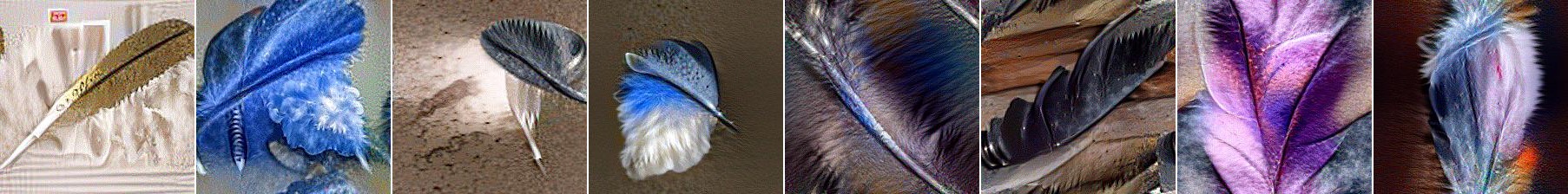}
\includegraphics[width=0.88\textwidth]{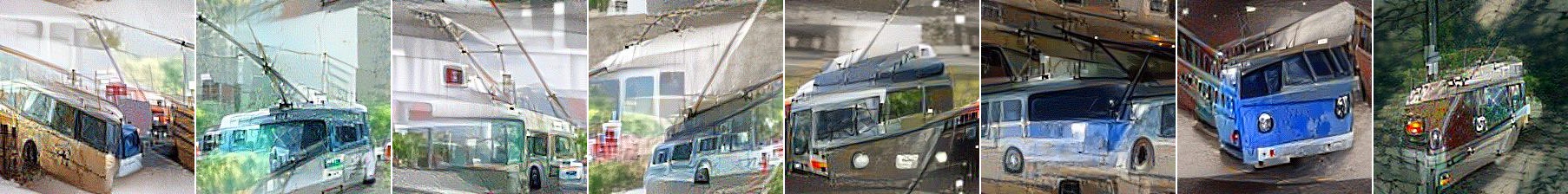}
\includegraphics[width=0.88\textwidth]{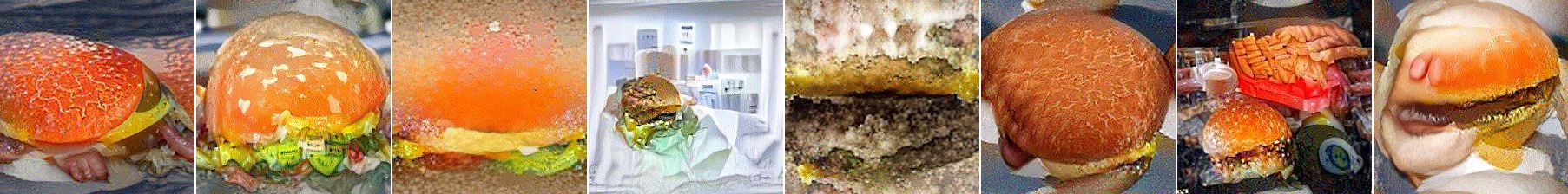}
\includegraphics[width=0.88\textwidth]{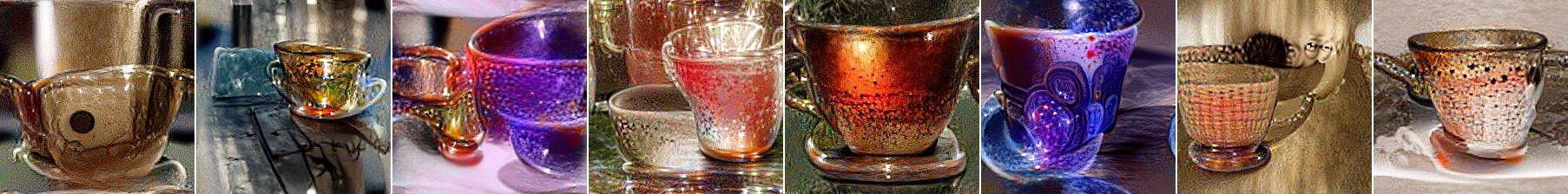}
\includegraphics[width=0.88\textwidth]{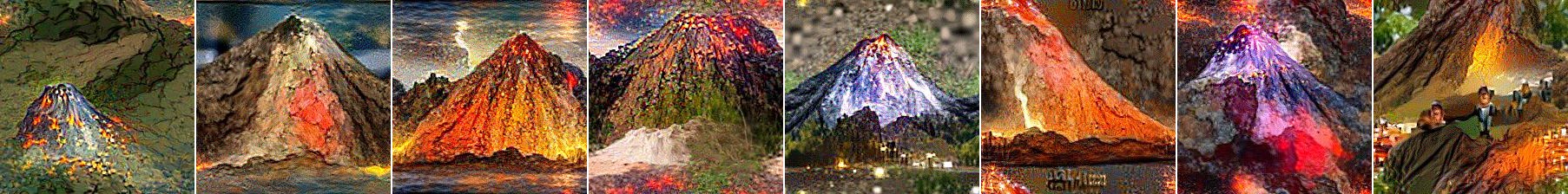}
\includegraphics[width=0.88\textwidth]{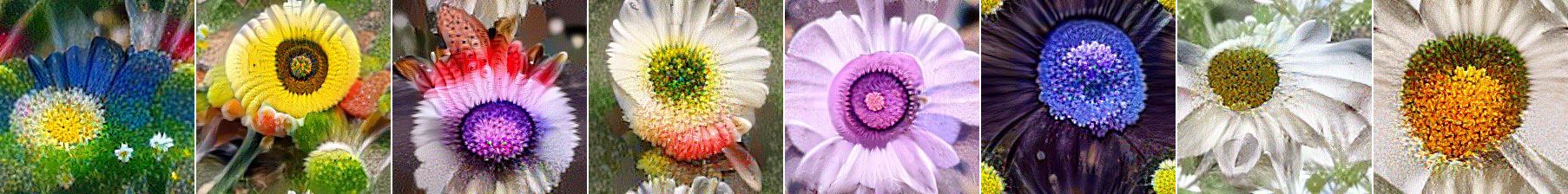}
\includegraphics[width=0.88\textwidth]{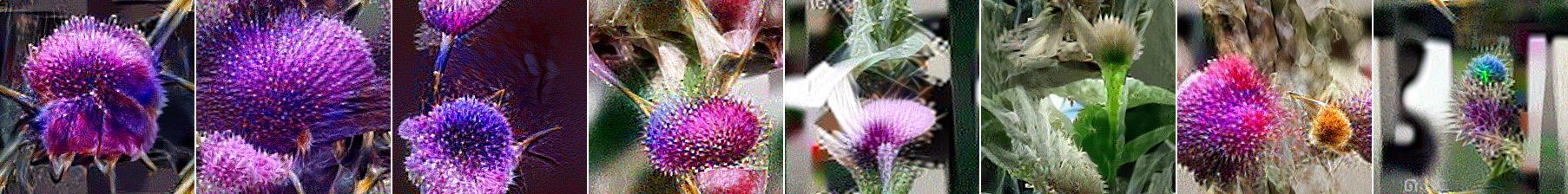}
\caption{Class-conditional $224\times224$ images obtained by DeepInversion given a ResNet50v1.5 classifier pretrained on ImageNet. Classes top to bottom: brown bear, quill, trolleybus, cheeseburger, cup, volcano, daisy, cardoon.}
\label{fig:resnetv1.5}
\end{figure*}

In this section, we show that we can distill information from a teacher network to a student network without using any real images at all. We apply DeepInversion to a ResNet50v1.5~\cite{resnet50v15url} trained on ImageNet to synthesize images. Using these images, we then train another randomly initialized ResNet50v1.5 from scratch. Below, we describe two practical considerations: a) image clipping, and b) multi-resolution synthesis, which we find can greatly help boost accuracy while reducing run-time. A set of images generated by DeepInversion on the pretrained ResNet50v1.5 is shown in Fig.~\ref{fig:resnetv1.5}. The images demonstrate high fidelity and diversity. 

\noindent\textbf{a) Image clipping.} 
We find that enforcing the synthesized images to conform to the mean and variance of the data preprocessing step helps improve accuracy. Note that commonly released networks use means and variances computed on ImageNet. We clip synthesized images to be in the $[-m / s, m / s ]$ range, with $m$ representing the per channel mean, and $s$ per channel standard deviation.

\noindent\textbf{b) Multi-resolution synthesis.} We find that we can speed up DeepInversion by employing a multi-resolution optimization scheme. We first optimize the input of resolution $112\times112$ for $2$k iterations. Then, we up-sample the image via nearest neighbor interpolation to $224\times224$, and then optimize for an additional $1$k iterations. This speeds up DeepInversion to $84$ images per $6$ minutes on an NVIDIA V100 GPU. Hyperparameters 
are given in the Appendix. 

\noindent\textbf{Knowledge transfer.} We synthesize 140k images via DeepInversion on ResNet50v1.5~\cite{resnet50v15url} to train a student network with the same architecture from scratch. Our teacher is a 
pretrained ResNet50v1.5 that achieves $77.26\%$ top-1 accuracy. We apply knowledge distillation for 90/250 epochs, with temperature $\tau=3$, initial learning rate of $1.024$, batch size of $1024$ split across eight V100 GPUs, and other settings the same as in the original setup~\cite{resnet50v15url}. Results are summarized in Table~\ref{tab:resnet50v15_scratch}. The proposed method, given only the trained ResNet50v1.5 model, can teach a new model from scratch to achieve a $73.8\%$ accuracy, which is only $3.46\%$ below the accuracy of the teacher.

\begin{table}[!t]
\centering
\resizebox{0.8\linewidth}{!}{
\midsepremove
\begin{tabular}{lccc}
\toprule
Image source    & Real images & Data amount          &Top-1 acc. \\ 
\midrule 
\midrule 
Base model       & \checkmark & 1.3M           &$77.26\%$ \\
\midrule
\rowcolor{gray}\multicolumn{4}{c}{Knowledge Transfer, 90 epochs} \\
\rowcolor{white} 
ImageNet         & \checkmark & 215k         &$76.1\%$       \\ 
MS COCO          & \checkmark & 127k         &$72.3\%$       \\
Generator, BigGAN & & 215k                &$64.0\%$ \\
DeepInversion (DI)& & 140k               &$68.0\%$ \\
\midrule 
\rowcolor{gray}\multicolumn{4}{c}{Knowledge Transfer, 250 epochs, with mixup} \\
\rowcolor{white} 
DeepInversion (DI)& & 140k              &$73.8\%$ \\
\bottomrule                                                   
\end{tabular}
}
\midsepdefault
\caption{Knowledge transfer from the trained ResNet50v1.5 to the same network initialized from scratch. }
\label{tab:resnet50v15_scratch}
\end{table}

\subsection{Application III: Data-free Continual Learning}
\label{sec:imagenet_incremental}
Data-free continual learning is another scenario that benefits from the image generated from DeepInversion. 
The main idea of continual learning is to add new classification classes to a pretrained model without comprehensive access to its original training data.
To the best of our knowledge, only LwF~\cite{li2017learning} and LwF.MC~\cite{rebuffi2017icarl} achieve data-free continual learning. Other methods require information that can only be obtained from the original dataset, \eg, a subset of data (iCaRL~\cite{rebuffi2017icarl}), parameter importance estimations (in the form of Fisher matrix in EWC~\cite{Kirkpatrick2016OvercomingCF}, contribution to loss change in SI~\cite{Zenke2017ContinualLT},
posterior of network weights in
VCL~\cite{nguyen2018variational}), or training data representation (encoder~\cite{RannenTrikiAljundi2017encoder}, GAN~\cite{Hu2019OvercomingCF, Shin2017ContinualLW}). Some methods rely on network modifications, \eg, Packnet~\cite{Mallya2017PackNetAM} and Piggyback~\cite{Mallya2018PiggybackAA}. In comparison, DeepInversion does not need network modifications or the original (meta-)data, as BN statistics are inherent to neural networks.

In the class-incremental setting, a network is initially trained on a dataset with classes $\mathcal{C}_o$, \eg, ImageNet~\cite{deng2009imagenet}. Given new class data $(x_k, y_k),~y_k \in \mathcal{C}_k$, \eg, from CUB~\cite{WelinderEtal2010}, the pretrained model is now required to make predictions in a combined output space $\mathcal{C}_o\cup\mathcal{C}_k$. Similar to prior work, we take a trained network \big(denoted $p_o(\cdot)$, effectively as a \textit{teacher}\big), make a copy \big(denoted $p_k(\cdot)$, effectively as a \textit{student}\big), and then add new randomly initialized neurons to $p_k(\cdot)$'s final layer to output logits for the new classes. We train $p_k(\cdot)$ to classify simultaneously over all classes, old and new, while network $p_o(\cdot)$ remains fixed.

\begin{table}[!t]
    \centering
    \begin{adjustbox}{width=.96\columnwidth}
    \setlength{\tabcolsep}{1pt}
    \midsepremove
    \begin{tabular}{lcccc}
    \toprule
    \multirow{2}{*}{Method} & \multicolumn{4}{c}{Top-1 acc. (\%) } \\
    & Combined & ImageNet & CUB & Flowers \\
    \midrule
    \rowcolor{gray}\multicolumn{5}{c}{ImageNet + CUB ($1000$ $\rightarrow$ $1200$ outputs)} \\
    LwF.MC~\cite{rebuffi2017icarl} & $47.64$ & $53.98$ & $41.30$ & -- \\
    DeepDream~\cite{mordvintsev2015deepdream} & $63.00$ & $56.02$ & $\mathbf{69.97}$ & -- \\
    DeepInversion (\textbf{Ours}) & $\mathbf{67.61}$ & $\mathbf{65.54}$ & $69.68$ & -- \\
    \midrule
    Oracle (distill) & $69.12$ & $68.09$ & $70.16$ & -- \\
    Oracle (classify) & $68.17$ & $67.18$ & $69.16$ & -- \\
    \midrule
    \rowcolor{gray}\multicolumn{5}{c}{ImageNet + Flowers ($1000$ $\rightarrow$ $1102$ outputs)} \\
    LwF.MC~\cite{rebuffi2017icarl} & $67.23$ & $55.62$ & -- & $78.84$ \\
    DeepDream~\cite{mordvintsev2015deepdream} & $79.84$ & $65.69$ & -- & $\mathbf{94.00}$ \\
    DeepInversion (\textbf{Ours}) & $\mathbf{80.85}$ & $\mathbf{68.03}$ & -- & $93.67$ \\
    \midrule
    Oracle (distill) & $80.71$ & $68.73$ & -- & $92.70$ \\
    Oracle (classify) & $79.42$ & $67.59$ & -- & $91.25$ \\
    \midrule
    \rowcolor{gray}\multicolumn{5}{c}{ImageNet + CUB + Flowers ($1000$ $\rightarrow$ $1200$ $\rightarrow$ $1302$ outputs)} \\
    LwF.MC~\cite{rebuffi2017icarl} & $41.72$ & $40.51$ & $26.63$ & $58.01$ \\
    DeepInversion (\textbf{Ours}) & $\mathbf{74.61}$ & $\mathbf{64.10}$ & $\mathbf{66.57}$ & $\mathbf{93.17}$ \\
    \midrule
    Oracle (distill) & $76.18$ & $67.16$ & $69.57$ & $91.82$ \\
    Oracle (classify) & $74.67$ & $66.25$ & $66.64$ & $91.14$ \\
    \bottomrule
    \end{tabular}
    \midsepdefault
    \end{adjustbox}
    \caption{Continual learning results that extend the network output space, adding new classes to ResNet-18. Accuracy over \emph{combined} classes $\mathcal C_o \cup\mathcal C_k$ reported on individual datasets. Average over datasets also shown (datasets treated equally regardless of size, hence ImageNet samples have less weight than CUB or Flowers samples).
    }
    \label{tab:incremental_learning}
    \vspace{-1mm}
\end{table}

\noindent\textbf{Continual learning loss.} 
We formulate a new loss with DeepInversion images as follows. We use same-sized batches of DeepInversion data $(\hat x, p_o(\hat x))$ and new class real data $(x_k, y_k)$ for each training iteration. For $\hat x$, we use the original model to compute its soft labels $p_o(\hat x)$, \textit{i.e.}, class probability among old classes, and then concatenate it with additional zeros as new class probabilities. We use a KL-divergence loss between predictions $p_o(\hat x)$ and $p_k(\hat x)$ on DeepInversion images for prior memory, and a cross-entropy (CE) loss between one-hot $y_k$ and prediction $p_k(x_k)$ on new class images for emerging knowledge. Similar to prior work~\cite{li2017learning,rebuffi2017icarl}, we also use a third KL-divergence term between the new class images' old class predictions $p_k(x_k|y\in\mathcal{C}_o)$ 
and their original model predictions $p_o(x_k)$. This forms the loss
\begin{eqnarray}
  \begin{aligned}
    \mathcal{L}_{\text{CL}} = & \text{KL}\big(p_o(\hat x), p_k(\hat x)\big) + \mathcal{L}_{xent}\big(y_k, p_k(x_k)\big) \\
    &+ \text{KL}\big(p_o(x_k|y\in\mathcal{C}_o), p_k(x_k|y\in\mathcal{C}_o)\big).
\end{aligned}
\label{eq:loss_incremental}
\end{eqnarray}
\vspace{-2mm}

\noindent\textbf{Evaluation results.} We add new classes from the CUB~\cite{WelinderEtal2010}, Flowers~\cite{nilsback2008automated}, and both CUB and Flowers datasets to a ResNet-18~\cite{he2016deep} classifier trained on ImageNet~\cite{deng2009imagenet}. Prior to each step of addition of new classes, we generate 250 DeepInversion images per old category. We compare our results with prior class-incremental learning work LwF.MC~\cite{rebuffi2017icarl} as opposed to the task-incremental LwF~\cite{li2017learning} that cannot distinguish between old and new classes. We further compare results with oracle methods that break the data-free constraint: we use the same number of real images from old datasets in place of $\hat x$, with either their ground truth for classification loss or their soft labels from $p_o(\cdot)$ for KL-divergence distillation loss. The third KL-divergence term in Eq.~\ref{eq:loss_incremental} is omitted in this case. Details are given in the Appendix.  

Results are shown in Table~\ref{tab:incremental_learning}. Our method significantly outperforms LwF.MC in all cases and leads to consistent performance improvements over DeepDream in most scenarios. Our results are very close to the oracles (and occasionally outperform them), showing DeepInversion's efficacy in replacing ImageNet images for continual learning. We verify results on VGG-16 and discuss limitations in the Appendix. 
\section{Discussion}
We next provide additional discussions on data-free quantization, and the limitations of the proposed method. \\

\noindent\textbf{Data-free quantization.} While we have studied three data-free tasks in this work, the proposed paradigm of data-free knowledge distillation via model inversion easily scales to other applications, such as the task of data-free network quantization as independently studied in~\cite{haroush2020knowledge, cai2020zeroq}. Haroush \textit{et al.}~\cite{haroush2020knowledge} explore \textit{The Knowledge Within} a trained network for inverted images towards the 4- and 8-bit quantization of ResNet-18, MobileNet V2, and DenseNet-121 networks. Cai~\textit{et al.}~\cite{cai2020zeroq} propose the ZeroQ framework based on only inverted images and knowledge distillation for data-free and zero-shot quantization. ZeroQ demonstrates less than 0.2\% accuracy loss when quantizing networks such as ResNets, MobileNet V2, Inception, SqueezeNets, etc., over MS COCO and ImageNet datasets. Both methods lead to efficient quantized models without the need of original data or any natural images.\\

\noindent\textbf{Limitations.} We discuss the limitations of the proposed approach as follows:
\begin{itemize}[leftmargin=*,noitemsep,topsep=0pt]
\setlength{\parskip}{0pt}
\setlength{\parsep}{1pt} 
  \item \textbf{Image synthesis time.}
  Generating 215K ImageNet samples of 224$\times$224 resolution for a ResNet-50 takes 2.8K NVIDIA V100 GPU-hours, or 22 hours on 128 GPUs. This time scales linearly with the number of synthesized images. The multi-resolution scheme described in Section~\ref{sec:imagenet_teaching} reduces this time by 10.7$\times$ (0.26K GPU-hours / 4 hours on 64 GPUs).
 
 \item \textbf{Image color and background similarity.} 
 We believe there are two possible reasons for this similarity. 1) The method uncovers and visualizes the unique discriminative characteristics of a CNN classifier, which can guide future work on neural network understanding and interpretation. Post-training, the network learns to capture only the informative visual representations to make a correct classification. For example, the key features of a target object are retained, \eg, detailed bear heads in Fig.~\ref{fig:resnetv1.5} or the fur color/patterns of penguins and birds in Fig.~\ref{fig:imagenet_inversion}, whereas the background information is mostly simplified, \eg, green for grass or blue for ocean.  
2) For all the images synthesized in this work, we use a default Gaussian distribution with zero mean and unit variance to initialize all the pixels, which may lead to unimodal behaviors. We have also observed that the style varies with the choice of the optimization hyperparameters. 

 \item \textbf{Continual learning class similarity.} We implemented DeepInversion on iCIFAR and iILSVRC~\cite{rebuffi2017icarl} (two splits) and observed statistically equivalent or slightly worse performance compared to LwF.MC. We suspect that the synthesized images are more effective in replacing old class images that are \emph{different} from the new images, compared to a case where the old and new images are similar (\eg, random subsets of a pool of classes).

\end{itemize}

\section*{Conclusions}
We have proposed DeepInversion for synthesizing training images with high resolution and fidelity given just a trained CNN. Further, by using a student-in-the-loop, our Adaptive DeepInversion method improves image diversity. Through extensive experiments, we have shown that our methods are generalizable, effective, and empower a new class of ``data-free'' tasks of immense practical significance. 

\section*{Acknowledgments}

We would like to thank Arash Vahdat, Ming-Yu Liu, and Shalini De Mello for in-depth discussions and comments.


{\small
\bibliographystyle{ieee}
\bibliography{bib}
}

\clearpage
\appendix
\section*{Appendix}

We provide more experimental details in the following sections. First, we elaborate on CIFAR-10 experiments, followed by additional details on ImageNet results. We then give details of experiments on data-free pruning (Section~\ref{sec:imagenet_pruning} of the main paper), data-free knowledge transfer (Section~\ref{sec:imagenet_teaching} of the main paper), and data-free continual learning (Section~\ref{sec:imagenet_incremental} of the main paper). 

\section{CIFAR-10 Appendix}
\subsection{Implementation Details}
We run each knowledge distillation experiment between the teacher and student networks for $250$ epochs in all, with an initial learning rate of $0.1$, decayed every $100$ epochs with a multiplier of $0.1$. One epoch corresponds to $195$ gradient updates. Image generation runs 4 times per epoch, in steps of $50$ batches when VGG-11-BN is used as a teacher, and 2 times per epoch for ResNet-34. The synthesized image batches are immediately used for network update (the instantaneous accuracy on these batches is shown in Fig.~\ref{fig:cifar10_training}) and are added to previously generated batches. In between image generation steps, we randomly select previously generated batches for training. 

\subsection{BatchNorm vs. Set of Images for $\mathcal{R}_\text{feature}$}
DeepInversion approximates feature statistics $\mathbb{E}\big(\mu_{l}(x) | \mathcal{X}\big)$ and $\mathbb{E}\big(\sigma^2_{l}(x) | \mathcal{X}\big)$ in $\mathcal{R}_\text{feature}$ (Eq.~\ref{eqn:fmap}) using BN parameters. As an alternative, one may acquire the information by running a subset of original images through the network. We compare both approaches in Table~\ref{tab:bn_partial}. From the upper section of the table, we observe that feature statistics estimated from the image subset also support DeepInversion and Adaptive DeepInversion, hence advocate for another viable approach in the absence of BNs. It only requires 100 images to compute feature statistics for Adaptive DeepInversion to achieve almost the same accuracy as with BN statistics.

\begin{table}[h]
\centering
\resizebox{0.8\linewidth}{!}{
\begin{tabular}{lcc}
\toprule
    \# Samples              & DI                            &ADI \\
                            & Top-1 acc. (\%)               &  Top-1 acc. (\%)   \\\midrule
    $1$                     & $61.78$                      & $84.28$ \\
    $10$                    & $80.94$                      & $89.99$ \\
    $100$                   & $83.64$                      & $90.52$ \\
    $1000$                  & $84.53$                      & $90.62$ \\\midrule
    BN statistics    & $84.44$                      & $90.68$ \\\bottomrule
\end{tabular}
}
\caption{CIFAR-10 ablations given mean and variance estimates based on (i) up: calculations from randomly sampled original images, and (ii) bottom: BN running\_mean and running\_var parameters. The teacher is a VGG-11-BN model at $92.34\%$ accuracy. The student is a freshly initialized VGG-11-BN. DI: DeepInversion; ADI: Adaptive DeepInversion.}
\label{tab:bn_partial}
\end{table}

\section{ImageNet Appendix}

\subsection{DeepInversion Implementation}
\label{sec:DI_implementation}
We provide additional implementation details of DeepInversion next. The total variance regularization $\mathcal{R}_{\text{TV}}$ in Eq.~\ref{eqn:r_prior} is based on the sum of $\ell_2$ norms between the base image and its shifted variants: (i) two diagonal shifts, (ii) one vertical shift, and (iii) one horizontal shift, all by one pixel. We apply random flipping and jitter ($<30$ pixels) on the input before each forward pass. We use the Adam optimizer with $\beta_1=0.9, \beta_2=0.999$ and $\epsilon=1\cdot10^{-8}$ given a constant learning rate of $0.05$. We speed up the training process using 
half-precision floating point (FP16) via the APEX library.

\section{Data-free Pruning Appendix}

\subsection{Hardware-aware Loss}
Our proposed pruning criterion considers actual latency on the target hardware for more efficient pruning. Characterized by Eq.~\ref{eq:latency_aware_loss}, in iterative manner neurons are ranked according to $\mathcal{I}_\mathcal{S}  (\textbf{W})$ and the least important ones are removed. The new scheme leverages $\mathcal{I}_{\mathcal{S}, err}$ and $ \mathcal{I}_{\mathcal{S}, lat}$ to focus on absolute changes in network error and inference latency, specifically, when the filter group $s\in\mathcal{S}$ is zeroed out from the set of neural network parameters $\textbf{W}$. 

Akin to ~\cite{molchanov2019importance}, we approximate $\mathcal{I}_{\mathcal{S}, err}$ as the sum of the first-order Taylor expansion of individual filters 
\begin{equation}
    \mathcal{I}_{\mathcal{S}, err} (\textbf{W}) \approx \sum_{s \in \mathcal{S}}\mathcal{I}_s^{(1)} (\textbf{W}) = \sum_{s \in \mathcal{S}}(g_s w_s)^2,
\end{equation}
where $g_s$ and $w_s$ denote the gradient and parameters of a filter $s$, respectively. We implement this approximation via gate layers, as mentioned in the original paper. 

The importance of a group of filters in reducing latency can be assessed by removing it and obtaining the latency change
\begin{equation}
    \mathcal{I}_{\mathcal{S}, lat} = \text{LAT}(\textbf{W}|w_s=0, s \in \mathcal{S}) - \text{LAT}(\textbf{W}),
\end{equation}
where $\text{LAT}(\cdot)$ denotes the latency of an intermediate pruned model on the target hardware. 

Since the vast majority of computation stems from convolutional operators, we approximate the overall network latency as the sum of their run-times. This is generally appropriate for inference platforms like mobile GPU, DSP, and server GPU~\cite{chamnet, fbnet}. We model the overall latency of a network as: 
\begin{equation}
\label{eqn:latency}
    \text{LAT}(\textbf{W}) = \sum_l \text{LAT} (o_l^{(\textbf{W})}),
\end{equation}
where $o_l$ refers to the operator at layer $l$. By benchmarking the run-time of each operator in hardware into a single look-up-table (LUT), we can easily estimate the latency of any intermediate model based on its remaining filters. The technique of using a LUT for latency estimation has also been studied in the context of neural architecture search (NAS)~\cite{chamnet, fbnet}. For pruning, the LUT consumes substantially less memory and profiling effort than NAS: instead of an entire architectural search space, it only needs to focus on the convolutional operations given \textit{reduced numbers} of input and output filters against the original operating points of the pre-trained model. 

\subsection{Implementation Details}
Our pruning setup on the ImageNet dataset follows the work in~\cite{molchanov2019importance}. For knowledge distillation given varying image sources, we experiment with temperature $\tau$ from $\{5, 10, 15\}$ for each pruning case and report the highest validation accuracy observed. We profile and store latency values in the LUT in the following format:
\begin{equation}
\label{eqn:conv_operators}
    key = [c_{in}, c_{out}, kernel^*, stride^*, fmap^*],
\end{equation}
where $c_{in}, c_{out}, kernel, stride,$ and $fmap$ denote input channel number, output channel number, kernel size, stride, and input feature map dimension of a Conv2D operator, respectively.
Parameters with superscript $^*$ remain at their default values in the teacher model. We scan input and output channels to cover all combinations of integer values that are \textit{less than} their initial values. Each key is individually profiled on a V100 GPU with a batch size 1 with CUDA 10.1 and cuDNN 7.6 over eight computation kernels, where the mean value of over $1000$ computations for the fastest kernel is stored. Latency estimation through Eq.~\ref{eqn:latency} demonstrates a high linear correlation with the real latency ($R^2=0.994$). For hardware-aware pruning, we use $\eta=0.01$ for Eq.~\ref{eq:latency_aware_loss} to balance the importance of $\mathcal{I}_{\mathcal{S}, err}$ and $ \mathcal{I}_{\mathcal{S}, lat}$, and prune away $32$ filters each time in a group size of $16$. We prune every $30$ mini-batches until the pre-defined filter/latency threshold is met, and continue to fine-tune the network after that. We use a batch size of $256$ for all our pruning experiments. To standardize input image dimensions, default ResNet pre-processing from PyTorch is applied to MS COCO and PASCAL VOC images. 

\subsection{Hardware-aware Loss Evaluation}
As an ablation, we show the effectiveness of the hardware-aware loss (Eq.~\ref{eq:latency_aware_loss} in Section~\ref{sec:imagenet_pruning}) by comparing it with the pruning baseline in Table~\ref{tab:hardware_aware_comparison}. We base this analysis on the ground truth data to compare with prior art. Given the same latency constraints, the proposed loss improves the top-1 accuracy by $0.5\%$-$14.8\%$. 

\begin{table}[h]
\centering
\resizebox{0.95\linewidth}{!}{
\begin{tabular}{lcc}
\toprule
V100 Lat.           & Taylor~\cite{molchanov2019importance}                 & Hardware-aware Taylor (\textbf{Ours}) \\
    (ms)                    & Top-1 acc. (\%)            &  Top-1 acc. (\%)   \\\midrule
    $4.90$ - baseline       & $76.1$                      & $76.1$ \\\midrule
    $4.78$                  & $76.0$                      & $76.5$ \\
    $4.24$                  & $74.9$                      & $75.9$ \\
    $4.00$                  & $73.2$                      & $73.8$ \\
    $3.63$                  & $69.2$                      & $71.6$ \\
    $3.33$                  & $55.2$                      & $70.0$ \\\bottomrule
\end{tabular}
}
\caption{ImageNet ResNet-50 pruning ablation with and without latency-aware loss given original data. Latency measured on V100 GPU at batch size 1. Top-1 accuracy corresponds to the highest validation accuracy for 15 training epochs. Learning rate is initialized to 0.01, decayed at the $10^{\text{th}}$ epoch. }
\label{tab:hardware_aware_comparison}
\end{table}

\subsection{Pruning without Labels}
Taylor expansion for pruning estimates the change in loss induced by feature map removal. Originally, it was proposed for CE loss given ground-truth labels of input images. We argue that the same expansion can be applied to the knowledge distillation loss, particularly the KL divergence loss, computed between the teacher and student output distributions. We also use original data in this ablation for a fair comparison with prior work and show the results in Table~\ref{tab:pruning_ablation}. We can see that utilizing KL loss leads to only $-0.7\%$ to $+0.1\%$ absolute top-1 accuracy changes compared to the original CE-based approach, while completely removing the need for labels for Taylor-based pruning estimates. 

\begin{table}[h]
\centering
\resizebox{0.9\linewidth}{!}{
\begin{tabular}{lcc}
\toprule
Filters pruned  & CE loss, w/ labels~\cite{molchanov2019importance} & KL Div., w/o labels (\textbf{Ours}) \\
    (\%)                    & Top-1 acc. (\%)             & Top-1 acc. (\%)   \\\midrule
    $0$ - baseline          & $76.1$                      & $76.1$ \\\midrule
    $10$                    & $72.1$                      & $72.0$ \\
    $20$                    & $58.0$                      & $58.1$ \\
    $30$                    & $37.1$                      & $36.4$ \\
    $40$                    & $17.2$                      & $16.6$ \\\bottomrule
\end{tabular}
}
\caption{ImageNet ResNet-50 pruning ablation with and without labels given original images. CE: cross-entropy loss between predictions and ground truth labels; KL Div: KullBack-Leibler divergence loss between teacher-student output distributions. Learning rate is 0, hence no fine-tuning is done.}
\label{tab:pruning_ablation}
\end{table}

\subsection{Distribution Coverage Expansion}
\label{sup:distribution}
\begin{figure}[h]
\centering
\includegraphics[width=0.45\textwidth]{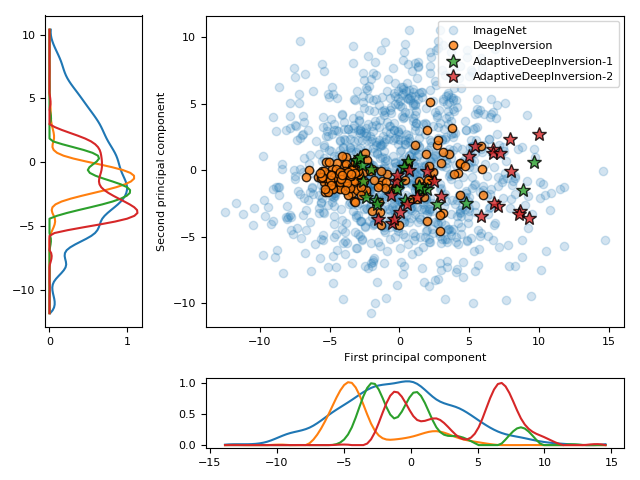}
\caption{Projection onto the first two principal components of the ResNet-50-avgpool feature vectors of ImageNet class `hand-held computer' training images. ADI-1/2 refers to additional images from round1/2 competition.}
\label{fig:pca_competition}
\end{figure}

Adaptive DeepInversion aims at expanding the distribution coverage of the generated images in the feature space through competition between the teacher and the student networks. Results of its impact are illustrated in Fig.~\ref{fig:pca_competition}. As expected, the distribution coverage gradually expands, given the two sequential rounds of competition following the initial round of DeepInversion.From the two side bars in Fig.~\ref{fig:pca_competition}, we observe varying ranges and peaks after projection onto each principal component from the three image generation rounds. 

\begin{figure}[t]
\centering

\subfigure[DeepInversion]{
\resizebox{0.6\linewidth}{!}{
\begin{tabular}{rccc}
\rotatebox{90}{\bf Inverted} & \includegraphics[width=.2\linewidth]{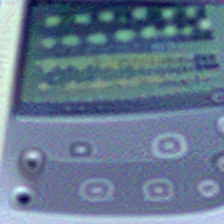} & \includegraphics[width=.2\linewidth]{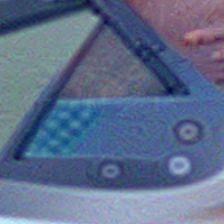} & \includegraphics[width=.2\linewidth]{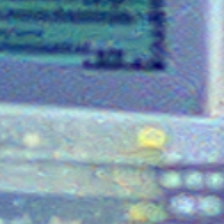}
\\
\bottomrule\\[-6pt]
\rotatebox{90}{\bf Closest real} & \includegraphics[width=.2\linewidth]{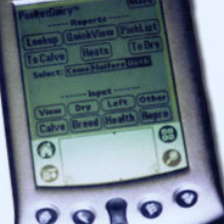} &
\includegraphics[width=.2\linewidth]{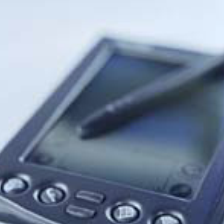} &
\includegraphics[width=.2\linewidth]{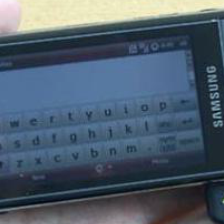}
\end{tabular}
}
}%
\subfigure[ADI]{
\resizebox{0.2\linewidth}{!}{
\begin{tabular}{c}
 \includegraphics[width=.2\linewidth]{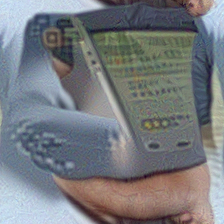} 
\\ 
\bottomrule\\[-6pt]
\includegraphics[width=.2\linewidth]{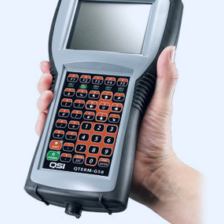} 
\end{tabular}
}
}

\caption{Nearest neighbors of the synthesized images in the ResNet-50-avgpool feature space for the ImageNet class `handheld computer' (a) without and (b) with the proposed competition scheme.
}
\label{fig:imagenet_competition}
\end{figure}

To further visualize the diversity increase due to competition loss (Eq.~\ref{eq:competition}), we compare the class of handheld computers generated with and without the competition scheme in Fig.~\ref{fig:imagenet_competition}. As learning continues, competition leads to the discovery of features for hands from the teacher's knowledge scope to challenge the student network. 
Moreover, generated images differ from their nearest neighbors, showing the efficacy of the approach in capturing distribution as opposed to memorizing inputs.  

\section{Data-free Knowledge Transfer Appendix}
We use the following parameters for the experiment on ResNet50v1.5: $\alpha_{\text{tv}}=1\cdot 10^{-4}$, $\alpha_{f}=0.01$, and a learning rate of $0.2$ for Adam. We generate images with an equal probability between the (i) multi-resolution scheme and (ii) the scheme described in Section~\ref{sec:imagenet_deepinversion} with $2$k iterations only to further improve image diversity. We clip the synthesized image $\hat{x}$ using
\begin{eqnarray}
    \hat{x} & = & \min\big(\max(\hat{x}, -m / s ), (1 - m) / s\big),
\end{eqnarray}
where $m=\{0.485, 0.456, 0.406\}$ and $s=$ $\{0.229,$ $ 0.224, $ $0.225\}$.

\section{Data-free Continual Learning Appendix}

\subsection{Implementation Details}
Our DeepInversion setup for this application follows the descriptions in Section~\ref{sec:imagenet_deepinversion} and Appendix~\ref{sec:DI_implementation} with minor modifications as follows. We use DeepInversion to generate \{250, 64\} images of resolution $224\times224$ per existing class in the pretrained \{ResNet-18, VGG-16-BN\}. These images are generated afresh after adding each new dataset. For \{ResNet-18, VGG-16-BN\}, we use a learning rate of $\{0.2,~0.5\}$, optimize for $10$k gradient updates in all, and decay the learning rate every $1.5$k steps with a $0.3$ multiplier. We use both $\ell_{2}$ and $\ell_1$ norms for total variance regularization at $\alpha_{\text{tv,}\ell_2}=\{3\cdot10^{-5},~6\cdot10^{-5}\},~\alpha_{\text{tv,}\ell_1}=\{1\cdot10^{-1}, 2\cdot10^{-1}\}$, jointly with $\alpha_{\ell_2}=0$ and $\alpha_{f}=\{1\cdot10^{-1},~3\cdot10^{-2}\}$ for DeepInversion. These parameters are chosen such that all loss terms are of the same magnitude, and adjusted to optimize qualitative results. 

Each method and dataset combination has individually-tuned learning rate and number of epochs obtained on a validation split using grid search, by optimizing the new dataset's performance while using the smallest learning rate and number of epochs possible to achieve 
this optimal performance. For each iteration, we use a batch of DeepInversion data $(\hat x, y_o(\hat x))$ and a batch of new class real data $(x_k, y_k)$. The batch size is $128$ for both kinds of data when training ResNet-18, and $64$ for VGG-16-BN. Similar to prior work~\cite{rebuffi2017icarl}, we reduce the learning rate to $20\%$ at $\frac{1}{3}$, $\frac{1}{2}$, $\frac{2}{3}$, and $\frac{5}{6}$ of the total number of epochs. We use stochastic gradient descent (SGD) with a momentum of $0.9$ as the optimizer. We clip the gradient $\ell_2$ magnitude to $0.1$, and disable all updates in the BN layers. Gradient clipping and freezing BN do not affect the baseline LwF.MC~\cite{rebuffi2017icarl} much ($\pm2\%$ change in combined accuracy after hyperparameter tuning), but significantly improve the accuracy of our methods and the oracles. We start with the pretrained ImageNet models provided by PyTorch. LwF.MC~\cite{rebuffi2017icarl} needs to use binary CE loss. Hence, we fine-tuned the model on ImageNet using binary CE with a small learning rate. The resulting ImageNet model is within $\pm 0.5\%$ top-1 error of the original model. We did not investigate the effect of the number of images synthesized on the performance.

\subsection{VGG-16-BN Results}
We show our data-free continual learning results on the VGG-16-BN network in Table~\ref{tab:incremental_learning_vgg}. The proposed method outperforms prior art~\cite{rebuffi2017icarl} by a large margin by enabling up to $42.6\%$ absolute increase in the top-1 accuracy. We observe a small gap of $<2\%$ combined error between our proposal and the best-performing oracle for this experimental setting, again showing DeepInversion's efficacy in replacing ImageNet images for the continual learning task. 

\begin{table}[h!]
    \centering
    \begin{adjustbox}{max width=0.91\columnwidth}
    \setlength{\tabcolsep}{3pt}
    \midsepremove
    \begin{tabular}{lcccc}
    \toprule
    \multirow{2}{*}{Method} & \multicolumn{4}{c}{Top-1 acc. (\%) } \\
    & Combined & ImageNet & CUB & Flowers \\
    \midrule
    \rowcolor{gray}\multicolumn{5}{c}{ImageNet + CUB ($1000$ $\rightarrow$ $1200$ outputs)} \\
    LwF.MC~\cite{rebuffi2017icarl} & $47.43$ & $64.38$ & $30.48$ & -- \\
    DeepInversion (\textbf{Ours}) & $\bf{70.72}$ & $\bf{68.35}$ & $\bf{73.09}$  & -- \\
    \midrule
    Oracle (distill) & $72.71$ & $71.95$ & $73.47$ & -- \\
    Oracle (classify) & $72.03$ & $71.20$ & $72.85$ & -- \\
    \midrule
    \rowcolor{gray}\multicolumn{5}{c}{ImageNet + Flowers ($1000$ $\rightarrow$ $1102$ outputs)} \\
    LwF.MC~\cite{rebuffi2017icarl} & $67.67$ & $65.10$ & -- & $70.24$ \\
    DeepInversion (\textbf{Ours}) & $\bf{82.47}$ & $\bf{72.11}$ & -- & $\bf{92.83}$ \\
    \midrule
    Oracle (distill) & $83.07$ & $72.84$ & -- & $93.30$ \\
    Oracle (classify) & $81.56$ & $71.97$ & -- & $91.15$ \\
    \bottomrule
    \end{tabular}
    \midsepdefault
    \end{adjustbox}
    \caption{Results on incrementally extending the network softmax output space by adding classes from a new dataset. All results are obtained using VGG-16-BN. }
    \label{tab:incremental_learning_vgg}
\end{table}

\subsection{Use Case and Novelty}
The most significant departure from prior work such as EWC~\cite{Kirkpatrick2016OvercomingCF} is that our DeepInversion-based continual learning can operate on \emph{any} regularly-trained model, given the widespread usage of BN layers. Our method eliminates the need for any collaboration from the model provider, even when the model provider (1) is unwilling to share any data, (2) is reluctant to train specialized models for continual learning, or (3) does not have the know-how to support a downstream continual learning application. This gives machine learning practitioners more freedom and expands their options when adapting existing models to new usage scenarios, especially when data access is restricted.

\end{document}